\documentclass[11pt]{article}
\usepackage[margin=1in]{geometry}
\usepackage{amsmath,amssymb,amsthm}
\usepackage{graphicx}
\usepackage{booktabs}
\usepackage{array}
\usepackage{multirow}
\usepackage{authblk}
\usepackage[round,authoryear]{natbib}
\usepackage{microtype}
\usepackage[
    colorlinks=true,
    citecolor=blue,
    linkcolor=blue,
    urlcolor=blue
]{hyperref}
\usepackage{longtable}
\usepackage{subcaption}
\usepackage{float}
\usepackage{appendix}
\usepackage{listings}
\title{\textbf{Quantifying Organizational Environmental Action
from Web Data and Large Language Models}}

\author[1]{Quinn Reynolds}
\author[2]{Daniel Shore}
\author[1,2]{Vianey Leos Barajas}
\author[2]{Tanhum Yoreh}
\author[1,2]{Meredith Franklin\thanks{Corresponding author:
\href{mailto:meredith.franklin@utoronto.ca}
{meredith.franklin@utoronto.ca}}}

\affil[1]{Department of Statistical Sciences, University of Toronto}
\affil[2]{School of the Environment, University of Toronto}

\date{}

\begin{document}

\maketitle


\begin{abstract}
Quantifying organizational environmental action from publicly available web content remains a challenging environmental data science problem because relevant information can be dispersed across multiple webpages and is primarily communicated through unstructured text. We present a scalable computational framework for transforming organizational web content into structured measures of environmental action and demonstrate the approach using Jewish congregations in the United States. We constructed a national database of 4,964 congregations by integrating multiple geospatial, knowledge-base, directory, and manually reviewed sources. Of these, 2,657 had active websites that were successfully crawled, producing a corpus of 154,454 webpages. We compared three approaches for detecting environmental actions: keyword retrieval followed by large language model (LLM) classification, semantic vector retrieval followed by LLM classification, and direct LLM classification classification without preliminary retrieval. Agreement with an expert human reviewer was lowest for keyword retrieval ($\kappa$ = 0.26), higher for semantic vector retrieval ($\kappa$ = 0.42), and similar for direct LLM classification ($\kappa$ = 0.40). Although semantic retrieval achieved the highest agreement, its retrieval recall was 0.87, indicating loss of relevant content before classification. Applied to the complete corpus, direct LLM classification identified at least one environmental action at 1,398 congregations (52.6\%), providing greater congregation-level coverage than either retrieval-based approach. These results demonstrate that preliminary retrieval can reduce computational cost but may exclude relevant information before it reaches the classifier. The framework provides a reproducible approach for extracting organization-level environmental information from unstructured web content that can be adapted to other institutions.
\end{abstract}


\section{Introduction}
Many environmental research questions increasingly depend on integrating large volumes of heterogeneous, data from diverse and continuously evolving sources \citep{blair2019}. While environmental data typically include measurements from monitoring networks, remote sensing observations, geospatial data, modeled and reanalysis products, surveys, and administrative records, many forms of environmental action are documented primarily in text, such as organizational websites, reports, news releases, and social media \citep{ghermandi2023}. These textual sources contain valuable information about sustainability initiatives \citep{ilieva2018}, conservation activities, climate commitments, and environmental governance, but they remain difficult to analyze systematically because they lack standardized structure. Recent advances in natural language processing (NLP) and large language models (LLMs) offer new opportunities to transform unstructured textual information into structured environmental datasets suitable for scientific analysis \citep{zhao2026}.

Automatically extracting environmental information from web data presents several methodological challenges \citep{kulkarni2021}. Organizations vary substantially in how they communicate environmental activities, using diverse terminology, writing styles, and website architectures. Keyword searches alone often suffer from low precision and recall because environmental concepts may be expressed implicitly or through context-dependent language \citep{naskar2024}. Transformer-based language models have substantially expanded the capacity for context-sensitive semantic classification and information extraction beyond traditional dictionary-based approaches \citep{devlin2019}. These models have rapidly become useful tools for converting unstructured text into structured variables across scientific domains. However, applying language models at web scale creates a tradeoff between computational efficiency and information loss. Retrieving or filtering text before LLM classification can substantially reduce computational demands but may exclude relevant content before it is evaluated by the model.

Although recent advances in environmental text mining have demonstrated the value of extracting information from scientific publications, sustainability reports, news articles, and social media, less attention has been given to developing and validating reproducible pipelines for measuring environmental action directly from organizational websites at national scale \citep{farrell2024}. Organizational websites provide a rich but underutilized source of information on environmental initiatives because they document organization-specific activities, programs, and practices that are often unavailable in structured databases. 

Places of worship provide a useful testbed for developing and evaluating such methods. Religious organizations represent one of the world’s largest civil society networks and collectively own substantial land, buildings, and financial resources while influencing the environmental attitudes and behaviors of millions of individuals \citep{gardner2002, jenkins2017,EomNg2023, HitzhusenTucker2013}. Scholars have argued that many religious traditions have increasingly embraced environmental stewardship through the broader “greening of religion” movement \citep{gottlieb2006}. However, empirical evidence on the prevalence and forms of environmental action among individual congregations remains limited because these activities are rarely collected in centralized databases and are instead documented by individual congregations across a variety of formats. Consequently, systematic measurement of environmental action across large numbers of congregations remains difficult.

In this study, we present and evaluate a scalable computational framework for measuring organizational environmental action from publicly available websites. We demonstrate the framework using Jewish congregations in the U.S. Our workflow integrates automated organization and website discovery, large-scale web crawling and archiving, and LLM-based information extraction to transform unstructured web content into congregation-level measures of environmental action. We compare three approaches for detecting environmental actions: (i) keyword retrieval followed by LLM classification, (ii) semantic vector retrieval followed by LLM classification, and (iii) direct LLM classification without preliminary retrieval. We evaluate these approaches against expert human classification and assess tradeoffs among agreement, retrieval recall, computational cost, and congregation-level coverage. The selected approach is subsequently used to characterize the type and framing of environmental actions and examine their geographic and organizational variation.

Beyond the specific application to religious organizations, the proposed methodology is adaptable to other types of institutions whose environmental activities are communicated through publicly available web content. By integrating web-scale data acquisition, LLM-based information extraction, human validation, and structured environmental classification, this study provides a reproducible approach for converting unstructured organizational web content into analysis-ready environmental information.

\section{Methodology}

We developed a computational framework to identify and classify documented environmental actions from publicly available organizational website content and applied it to Jewish places of worship in the United States (U.S.). The unit of analysis is the individual place of worship, with environmental actions were measured from publicly available content published on each congregation’s website. We define environmentalism as concern for the protection of the natural world and operationalize organizational environmentalism through environmental actions undertaken or encouraged by an organization. We define environmental action broadly as organizational practices, programs, or activities with actual or intended environmental relevance \citep{caldwellShadesGreenEnvironmental2022}.

The empirical application focused on U.S.\ Jewish places of worship. We focused on the U.S.\ because it has the largest Jewish population outside of Israel, providing a large and heterogeneous population for national-scale analysis. In addition, congregation websites are predominantly available in English, enabling consistent text processing. Finally, large-scale empirical data on Jewish environmentalism within the U.S.\ remain limited \citep{tirosh-samuelsonJewishEnvironmentalismUnited2024}.

\subsection{Database Construction}

We assembled the analytic database using a combination of programmatic searches and manual review. We drew on five complementary sources: the OpenStreetMap Overpass API; Wikidata, queried using SPARQL coordinate and name-plus-state matching; Wikipedia state-level synagogue lists covering all 50 states and the District of Columbia; denominational and federation
directories; and the Google Places API. 

To improve coverage in densely populated areas, Google Places was queried using an adaptive quadtree decomposition over a $1^{\circ}\times1^{\circ}$ grid covering the contiguous U.S.\ Grid cells approaching the API result limit were recursively subdivided to recover congregations that might otherwise be omitted in high density areas. We supplemented these sources with manual searches of state-level synagogue lists and denominational directories. 

Records from all sources were merged and deduplicated using congregation names and geographic coordinates, producing a single canonical database of Jewish places of worship. The database was implemented in PostgreSQL with the PostGIS and pgvector extensions. For each congregation, we recorded its name, city, state, street address and geographic coordinates, website, and denominational affiliation.

\subsection{Web Crawling and Archiving}

We created a fixed corpus of publicly available website content for each congregation with an active website. Archiving website content prior to classification ensured that subsequent analyses were based on a consistent snapshot rather than on webpages that could change during data collection or model evaluation.

Each website was crawled using a breadth-first traversal implemented in Python with the BeautifulSoup HTML parser \citep{BeautifulSoupDocumentation}. Beginning at the website root, the crawler followed in-domain links and retrieved up to 150 pages per website. Crawling was performed using ten concurrent workers, with each worker operating independently across websites. Media files and dynamic pages were excluded. Crawl state was retained so that interrupted crawls could be resumed without duplicating previously retrieved pages.

To minimize potential burdens on website operators and respect restrictions on automated access, the crawler used a fixed 0.2 second delay between requests to the same website and did not attempt to circumvent technical measures that blocked automated traffic. Websites that prevented access through automated traffic controls were therefore excluded from automated content collection. These restrictions were particularly consequential for congregations using shared website platforms, where access controls could be applied across multiple hosted websites. For example, automated access was restricted to a number of synagogue websites hosted on ShulCloud.

Websites that could not be crawled successfully using the initial crawl settings were subsequently subjected to a slower recovery crawl. This procedure began with a 5 second request delay, doubling upon failure up to a max of 80 seconds. This approach recovered content from additional websites, particularly those sharing common hosting infrastructure. For example, it recovered a number of Chabad websites that shared the same underlying chabad.org infrastructure. 

Because the complete webpage corpus was retained rather than only text fragments containing keywords, the corpus could subsequently be reanalyzed using alternative retrieval and classification procedures. All subsequent classification and analysis were conducted on this fixed corpus, representing website content retrieved through June 1, 2026. 

\subsection{Environmental Action Classification}

We used a large language model (LLM) as a classifier for several information-extraction tasks. In an LLM-classifier framework \citep{zhengJudgingLLMasaJudgeMTBench2023}, a pretrained general-purpose language model is provided with a task-specific prompt specifying classification definitions and decision rules and returns a structured classification for each text input. This approach enabled us to convert unstructured organizational website content into structured variables for quantitative analysis. We implemented this framework using Google Gemini 3.1 Flash-Lite, accessed through the Google Gemini developer API. Complete prompts for all classification tasks are provided in Appendix \ref{appendixA}.

The LLM was used for three tasks in this study: (1) identifying and categorizing environmental actions in website content (see \ref{appendixA1}), (2) inferring denominational affiliation when it could not be determined directly from available metadata (see \ref{appendixA2}), and (3) classifying the framing of identified environmental actions (see \ref{appendixA3}) . 

For environmental action identification and categorization (task 1), we evaluated three approaches that differed in how website content was selected for LLM classification: (1) keyword retrieval followed by LLM classification, (2) vector similarity retrieval followed by LLM classification, and (3) direct LLM classification without a preceding retrieval filter. The first two approaches used a two-stage procedure in which a computationally inexpensive retrieval method first identified candidate text, followed by LLM classification of the retained content. These retrieval steps were intended to reduce the amount of text submitted to the LLM, thereby decreasing computational cost and processing time. The direct LLM approach instead classified website content without requiring it to pass a keyword- or similarity-based retrieval filter. All three approaches used the same LLM for classification, allowing us to evaluate the tradeoff between computational efficiency and the potential loss of environmental actions during candidate retrieval. We used zero-shot prompting for environmental action classification because the large number of action types and substantial imbalance between environmental action and non-action text made constructing a representative few-shot prompt impractical.

\subsubsection{Environmental Action Taxonomy}

We defined environmental actions using a taxonomy developed from the Greening Sacred Spaces program of Faith \& the Common Good (FCG) \citep{FCG}. The final taxonomy consisted of 76 environmental action types organized into nine overarching categories: Community; Spirituality and Worship; Kitchen; Waste; Energy; Operations and Maintenance; Environmental and Climate Justice; Water; and Other. The 76 action types represent specific forms of environmental engagement within these broader categories; for example, Energy includes actions related to renewable energy and energy conservation, while Spirituality and Worship includes environmentally themed religious services, study, prayer, and ritual. Illustrative examples of each category from a random sample of our database are provided in Table \ref{tab:action_categories}, and the complete taxonomy of the 76 environmental action types is provided in Appendix \ref{appendixB} Table \ref{tab:action_taxonomy}.

\begin{longtable}{p{0.42\linewidth}p{0.52\linewidth}}
\caption{Environmental action categories and illustrative examples.}
\label{tab:action_categories}\\
\hline
\textbf{Environmental Action Category} & \textbf{Illustrative Actions} \\
\hline
Spirituality \& Worship &
Reclaim Shabbat as a moment to be at one with nature.  \\
Community &
Care for the environment. Feed the hungry. Welcome the stranger. Teach. Cook. Play. Hug. Listen. Care. Be one of nearly 1,000 congregants who volunteer thousands of service hours every year. Register for Volunteer Projects...\\
Kitchen &
...chicken, Vegan, and Gluten-Free Entrée choice plus Sides\\
Waste &
In June we will be donating to the --- Shelter and will be collecting gently used or new clothing such as T-shirts, socks, shorts, and hats.\\
Energy &
We've determined that the Temple's roof would be a great place to install solar panels. It's flat and under little shade throughout the year. Panels would not only support the Temple with electricity, they might also serve as a source of income. \\
Operations \& Maintenance &
...radiators powered by a new, well-sealed gas system in the basement.  \\
Environmental \& Climate Justice &
Land Acknowledgment --- recognizes and honors the original inhabitants who first settled in the valley of the Kwinitekw River. --- acknowledges that we are on Nonotuck land.\\
Water &
Fix or find an alternative to the drain pipe from the east roof to the city storm sewer. One solution would be a rain garden.  \\
Other &
GREEN TEAM Support the mission of ---'s Green Team with a directed donation. Previous donations have been used to: build compost bins, put in the EV charger, and populate the garden. Click here to donate to the GREEN TEAM.\\%
\hline
\end{longtable}

When an environmental action was identified, the LLM assigned it to one of the 76 environmental action types. Each action type corresponded to one of the nine overarching environmental action categories. Identified actions were subsequently aggregated within each congregation to produce congregation-level counts by action type and environmental action category.

\subsubsection{Keyword Retrieval and LLM Classification}

We developed an environmental keyword dictionary based on the Greening Sacred Spaces checklist \citep{FCG} and supplemented it with Judaism-specific terminology. The dictionary contained 104 exact words or phrases and 39 additional word stems designed to capture multiple lexical variants (ex: conserv* to identify "conservation", "conserve", and "conservationist").  

We searched the extracted text of each archived webpage for each keyword independently. Keyword occurrences were used to identify candidate passages for subsequent LLM classification and were not themselves considered evidence of an environmental action. Keyword occurrences separated by more than 15 characters were treated as distinct candidate locations. For each candidate location, we extracted a text chunk extending up to 500 characters before and after the matched keyword. Where neighboring keyword windows overlapped, the intervening text was divided at its midpoint to avoid duplicating text across candidate chunks. 

Each candidate chunk was then submitted to the LLM for classification. The LLM determined whether the candidate text described an environmental action and, if so, assigned the action to one of the 76 action types defined in the environmental action taxonomy.

\subsubsection{Vector Similarity Retrieval and LLM Classification}

For the vector-based approach, we used semantic similarity rather than lexical keyword matching to identify candidate environmental content. The extracted text of each webpage was embedded using the transformer-based sentence encoder \texttt{all-MiniLM-L6-v2}. Embeddings were stored and queried using the pgvector extension for PostgreSQL. Each of the 76 environmental action types in the taxonomy was separately embedded to produce a set of semantic query vectors. For each webpage, we calculated the maximum cosine similarity between its embedding and the 76 action type embeddings.  

Webpages with a maximum cosine similarity score greater than 0.25 were retained for subsequent LLM classification. The threshold of 0.25 was selected during method development to balance retrieval precision and recall. Because retrieval was based on semantic similarity rather than exact lexical matching, this approach could identify candidate content expressed using terminology not included in the keyword dictionary. Retained pages were then divided into consecutive chunks of approximately 500 characters,  with chunk boundaries shifted to the nearest whitespace to avoid truncating words. Each chunk was submitted to the LLM for environmental action identification and categorization.

\subsubsection{Direct LLM Classification}

The direct LLM approach eliminated the candidate retrieval stage. All extracted webpage text was divided into consecutive chunks of approximately 500 characters, with chunk boundaries shifted to the nearest whitespace to avoid truncating words. Each chunk was then submitted directly to the LLM for environmental action identification and categorization. 

Because this approach did not require text to contain a predefined keyword or exceed a semantic-similarity threshold, all retrieved website content was eligible for classification. This allowed the classifier to identify environmental actions expressed using language that might not be captured by either retrieval method, at the cost of submitting substantially more text to the LLM. 

\subsubsection{Human Validation and Method Selection}

We evaluated the three environmental action identification and categorization approaches against classifications made by an expert human reviewer. The human classifications were treated as the reference standard for evaluating the automated methods.

The validation sample size was determined using a Monte Carlo power analysis for a hypothesis test of Cohen's $\kappa$. We tested a null hypothesis of $\kappa_0 = 0.20$ against an alternative hypothesis of $\kappa_a=0.40$, allowing the human reviewer and LLM classifier to have different positive classification rates, assumed to be 1\% and 4\%, respectively. We conducted 200,000 Monte Carlo simulations for each candidate sample size from 700 to 1,400 in increments of 50. A sample size of 1,200 text chunks provided estimated statistical power of 80\%. We therefore randomly sampled 1,200 text chunks for independent classification by the expert reviewer. This relatively large sample was required because environmental actions were expected to be rare relative to text containing no environmental action. 

Agreement between the human reviewer and each automated detection method was evaluated using both raw percentage agreement and Cohen’s $\kappa$. Raw agreement was calculated as the proportion of observations for which the human reviewer and automated method assigned the same classification. Because high agreement can occur by chance when one outcome category is substantially more prevalent than another, we used Cohen’s $\kappa$ as the primary chance-corrected measure of inter-rater agreement. Cohen’s $\kappa$ was calculated as $\kappa = (p_o-p_e)/(1-p_e)$
where $p_o$ is the observed proportion of agreement between the human reviewer and automated classifier, and $p_e$ is the proportion of agreement expected by chance given the marginal classification frequencies. A $\kappa = 1$ represents complete agreement, $\kappa=0$ represents agreement equivalent to that expected by chance, and $\kappa<0$ represents agreement lower than expected by chance.

For the two approaches that included a candidate retrieval stage, we also evaluated retrieval recall, defined as the proportion of observations identified as environmental action positive by the human reviewer that were retained by the retrieval procedure for subsequent LLM classification. This allowed errors introduced during candidate retrieval to be distinguished from errors introduced during LLM classification.

\subsection{Secondary Classification and Contextual Variables}

Denominational affiliation was included as an organizational characteristic because religious denominations may differ in theological traditions, institutional structures, and social worldviews that can influence environmental attitudes and behaviors \citep{smithWhatsEvangelicalGot2018a,ncheSpiritualFocusClimate2020}.
Previous research suggests that religious beliefs and denominational affiliation may be associated with environmentalism, although the magnitude and consistency of denominational differences remain debated \citep{orellanoInfluenceReligionSustainable2020, guth2023, taylorLynnWhiteJr2016, wuthnowRestructuringAmericanReligion1996}.

We grouped congregations into nine denominational categories: Reconstructionist, Reform, Jewish Renewal, Conservative, Non-Denominational Progressive, Humanistic, Non-Denominational Conservative, Orthodox, and Chabad. Denominational affiliation was assigned directly when it could be determined from the congregation name, website, or available metadata. For places of worship without an explicit denominational identifier, we used the LLM with a zero-shot classification prompt to infer denominational affiliation from available website content. Congregations for which denominational affiliation could not be resolved were excluded from denomination-specific analyses. 

LLM-based denomination classification was assessed against independent classifications made by an expert human reviewer, which were treated as the reference standard, and agreement was evaluated using raw percentage agreement and Cohen’s $\kappa$, as described above. 

In addition to identifying environmental actions, we classified the environmental framing of each identified action. We distinguished three framing categories: explicit, implicit, and embedded environmental actions.

Explicit environmental actions were those for which the congregation directly stated an environmental motivation or objective \citep{baugh2019,caldwellShadesGreenEnvironmental2022}. Implicit environmental actions were environmentally relevant activities for which an environmental motivation was not explicitly stated \citep{caldwellShadesGreenEnvironmental2022}. Embedded environmental actions were environmentally relevant practices integrated into religious life, theology, ritual, or routine congregational activity and not necessarily presented as discrete environmental initiatives \citep{baugh2019}.

Environmental framing was classified using the LLM with a few-shot prompt containing one labeled example from each of the three framing categories \citep{brown2020}. Few-shot prompting was used because the task involved only three categories and the examples provided balanced representation across the three classes. The framing classifier was validated against classifications made by an expert human reviewer, with the human classifications treated as the reference standard. Agreement was evaluated using raw percentage agreement and Cohen’s $\kappa$ as described above. 

Finally, we characterized the local political context of each congregation by spatially linking its geographic coordinates to its U.S.\ congressional district. Districts were classified according to whether the Democratic or Republican candidate won the district in the 2024 U.S.\ federal election. We then compared the distributions of explicit, implicit, and embedded environmental action counts between congregations located in districts won by Democratic and Republican candidates.

\subsection{Statistical Analysis}
Environmental action counts were discrete and right skewed, with a relatively small number of congregations having substantially larger action counts. Descriptive summaries included frequencies, percentages, means, medians, and interquartile ranges (IQR), as appropriate. Potentially extreme observations were identified using a standard z-score threshold of $|z|>3$ applied to $log(1+x)$-transformed environmental action counts, where $x$ represents the observed action count. This procedure was used to reduce the influence of unusually extreme observations on group comparisons. Observations identified by this procedure were removed.

We used nonparametric methods for group comparisons, and all statistical tests were two-sided with statistical significance defined as $p<0.05$. For comparisons between two independent groups, differences in environmental action counts were evaluated using the Mann–Whitney $U$ test. For comparisons among three or more independent groups, differences in environmental action counts were evaluated using the Kruskal–Wallis test. When the Kruskal–Wallis test indicated a statistically significant difference among groups, post-hoc pairwise comparisons were conducted using Dunn’s test. The Benjamini–Hochberg (BH) procedure was applied to the post-hoc $p$-values to control the false discovery rate associated with multiple comparisons. For comparisons among related measurements from the same congregations, differences in environmental action counts were evaluated using the Friedman test. When the Friedman test indicated a statistically significant difference among related groups, post-hoc pairwise comparisons were conducted using the Nemenyi test to determine which groups differed. 

All analyses were conducted using Python 3.11.5.

\begin{figure}[ht]
    \centering
    \includegraphics[width=0.99\linewidth]{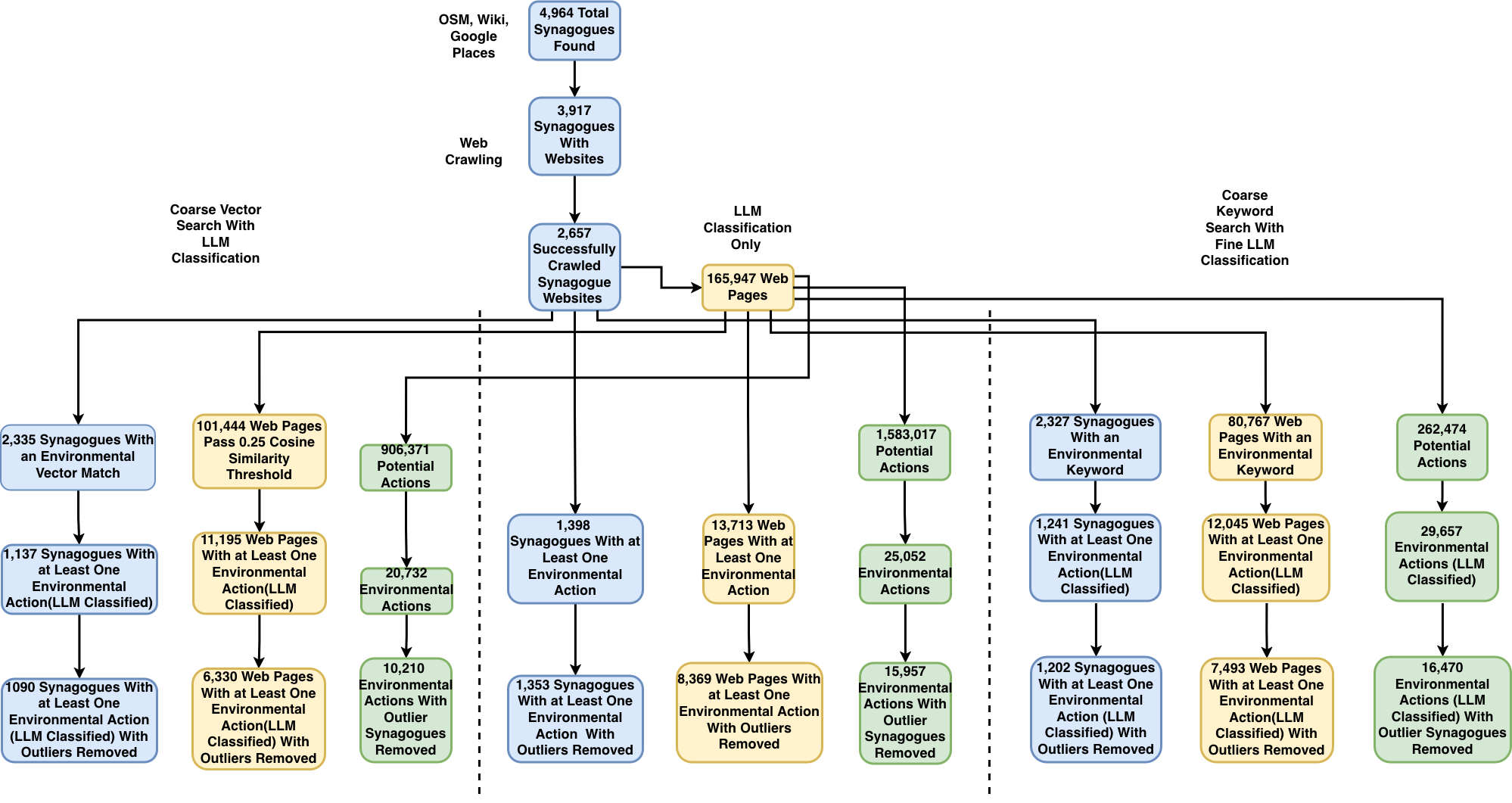}
    \caption{Flow diagram of study database creation, web crawling, and environmental action classification. }
    \label{fig:fig1}
\end{figure}

\section{Results}

\subsection{Study Population and Website Corpus}

The combined programmatic searches and manual review identified 4,964 Jewish congregations in the U.S.\ Of these, 3,917 had an identified website (Figure \ref{fig:fig1}). A total of 3,001 websites were reached by the crawler without a fatal error, and 2,657 yielded at least one retrievable webpage and therefore met the definition of an active website. Websites that could not be accessed included those that restricted automated traffic through technical access controls. The 2,657 congregations constituted the analytic population for the web-content analysis. Across these active websites, the crawler archived 158,454 webpages. 

\subsection{Environmental Action Classification Method Comparison and Validation}

The three environmental action detection approaches differed in their agreement with the validation sample of 1,200 randomly selected text chunks for human review (Table \ref{tab:tab1}). Keyword retrieval followed by LLM classification had the lowest agreement (Cohen’s $\kappa$ = 0.26; raw agreement = 82.1\%). Agreement was higher for vector retrieval followed by LLM classification ($\kappa$ = 0.42; raw agreement = 97.6\%) and direct LLM classification ($\kappa$ = 0.40; raw agreement = 96.7\%).

\begin{table}[ht]
\centering
\caption{Comparison of environmental action detection methods using expert human validation and computational performance (validation sample, $n=1{,}200$).}
\label{tab:tab1}
\tabcolsep=16pt%
\begin{tabular}{lccc}
\hline
& Keyword + LLM & Vector + LLM & Direct LLM \\
\hline
Candidate Chunks & 262,474  & 906,371 & 1,583,017 \\
Cohen's $\kappa$ &0.26  & 0.42 & 0.40 \\
Raw Agreement (\%) &82.1&97.6 &96.7\\
Retrieval Recall &0.94& 0.87& --\\
Cost (USD) & 28.77& 85.80 & 153.57\\
Time (hours) &1.1& 2.7  & 4.6 \\
\hline
\end{tabular}
\end{table}

Although the vector-based approach produced slightly higher agreement with the human reviewer than direct LLM classification, the vector retrieval stage had a recall of 0.87, indicating that some human-identified environmental content was excluded before it could be evaluated by the LLM. Keyword retrieval similarly had imperfect retrieval recall (0.94), demonstrating that both preliminary filtering approaches excluded some relevant content. For example, one congregation described a recently installed electric-vehicle charging station with the headline "At This Synagogue, You Can Charge Your Car While You Charge Your Soul." Although the underlying action, the installation of an electric-vehicle charger, was environmentally relevant, the unconventional description illustrates how organizational websites can describe environmental activities using language that may not be captured by predefined lexical or semantic retrieval criteria. Direct LLM classification avoids this source of information loss by evaluating website content without requiring it to first pass a retrieval filter. Thus, agreement with expert classification alone did not capture information lost during the preceding retrieval stage.

The increased coverage of direct LLM classification came at a higher computational cost (Table \ref{tab:tab1}). Direct LLM classification evaluated 1,583,017 text chunks, cost \$153.57 and required an estimated 4.6 hours of processing, compared with \$85.80 and 2.7 hours for vector retrieval followed by LLM classification and \$28.77 and 1.1 hours for keyword retrieval followed by LLM classification. The retrieval-based approaches therefore reduced computational demands by limiting the amount of website content submitted to the LLM, but at the cost of lower retrieval recall.

Differences between the detection approaches were also apparent when applied to the complete website corpus. Before application of downstream analytic exclusion criteria, keyword retrieval followed by LLM classification identified at least one environmental action at 1,241 congregations, vector retrieval followed by LLM classification identified at least one action at 1,137 congregations, and direct LLM classification identified at least one action at 1,398 congregations. Direct LLM classification therefore identified environmental activity at 261 additional congregations compared with the vector-based approach and 157 additional congregations compared with the keyword-based approach. Despite its greater computational cost, direct LLM classification was selected for subsequent analyses because it provided the greatest congregation-level coverage and avoided the loss of relevant content introduced by preliminary retrieval.

\subsection{Secondary Classification Validation}

The LLM used to classify congregational denominational affiliation and framing of environmental actions identified by the direct LLM approach showed high agreement with the human reviewer (Table \ref{tab:tab2}). The validation sample sizes of 20 congregations for denominational classification and 106 actions for environmental framing were determined via the Monte Carlo method to provide 80\% statistical power.

\begin{table}[ht]
\centering
\caption{Agreement between expert human and LLM classification for denominational affiliation and framing.}
\label{tab:tab2}
\tabcolsep=8pt%
\begin{tabular}{lcc}
\hline
& Denominational Classification & Environmental Action Framing \\ 
\hline
Number of Classes & 10 & 3 \\
Validation Sample Size ($n$) & 20 & 106 \\
Cohen's $\kappa$ & 1.00 &0.70 \\
Raw Agreement (\%) & 100.0 & 91.5\\
\hline
\end{tabular}
\end{table}

Classification into the 10 denomination classes showed complete agreement, with 100\% raw agreement and Cohen's $\kappa=1$ across 20 congregations in the validation sample. Environmental action framing, which classified actions as explicit, implicit or embedded, achieved 91.5\% raw agreement and Cohen's $\kappa = 0.70$ across the 106 validated actions.
These validation results supported the use of both secondary LLM classifications in subsequent analyses.

\subsection{Distribution of Environmental Actions}

\begin{figure}[H]
    \centering
    \includegraphics[width=0.75\linewidth]{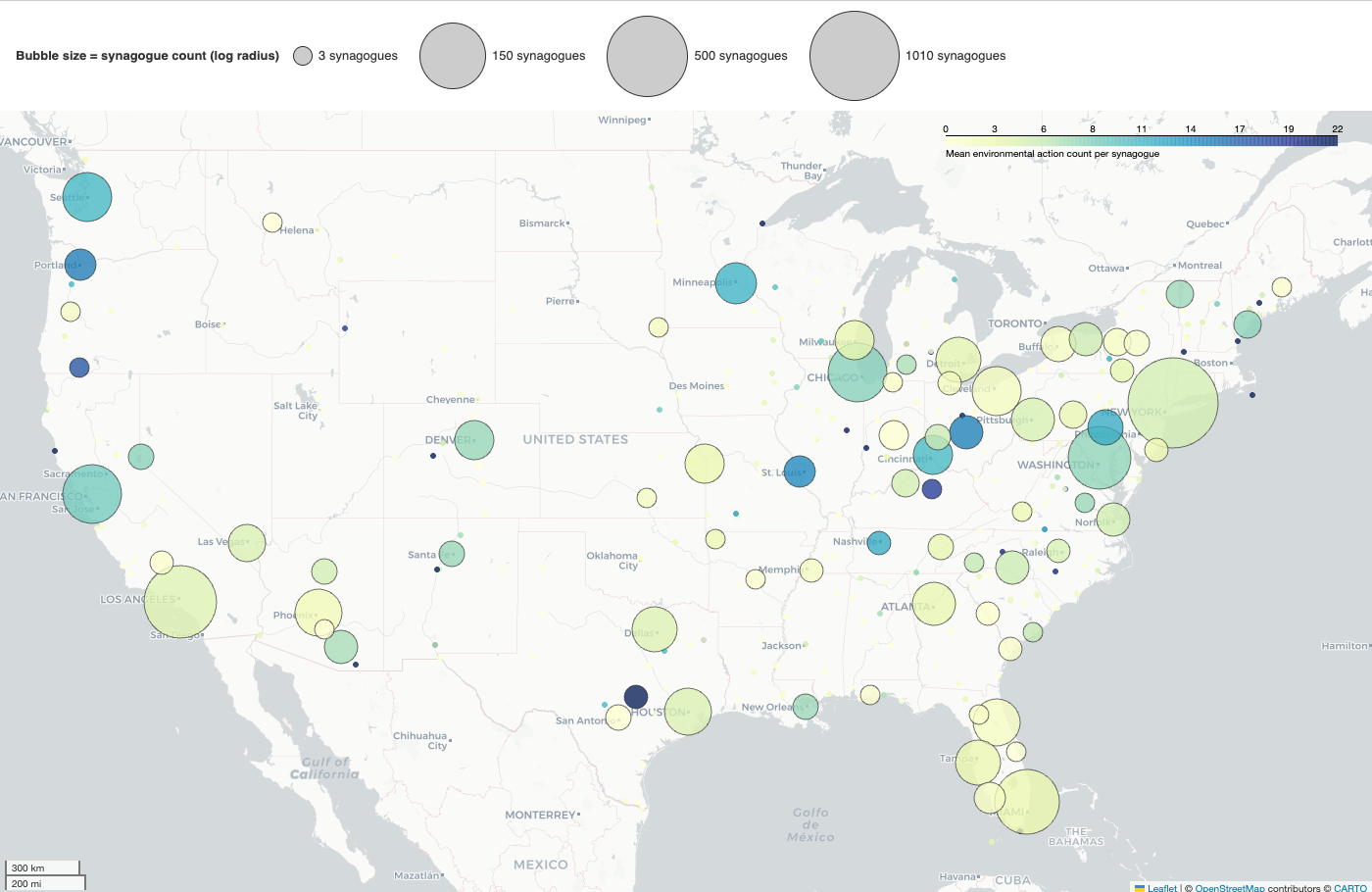}
    \caption{Geographic distribution of environmental action among U.S.\ Jewish congregations. Bubble size represents the number of congregations, colour represents the mean number of identified environmental actions ($\ge 1$) per congregation.}
    \label{fig:map}
\end{figure}

Across all 2,657 congregations with active websites, the mean number of identified environmental actions per congregation was 5.1. The distribution was strongly right-skewed, as the median was one action with interquartile range [IQR] 0–4. Among the 1,353 congregations with at least one identified environmental action, the mean increased to 10.0 actions per congregation. The maximum observed action count following application of the outlier exclusion criterion was 113. Environmental action counts also varied geographically across the U.S. (Figure \ref{fig:map}) 

The cluster centered on New York contained the largest number of congregations ($n=1,007$) with a mean of 5.6 identified environmental actions per congregation. Among the ten largest geographic clusters, Seattle  ($n=34$) had the highest mean number of environmental actions, with 13.7 environmental actions per congregation on average, whereas the Daytona Beach cluster ($n = 30$) had the lowest, with 1.6 environmental actions per congregation on average. 

\begin{figure}[H]
    \centering
        \begin{subfigure}{0.85\linewidth}
        \centering
    \includegraphics[width=\linewidth]{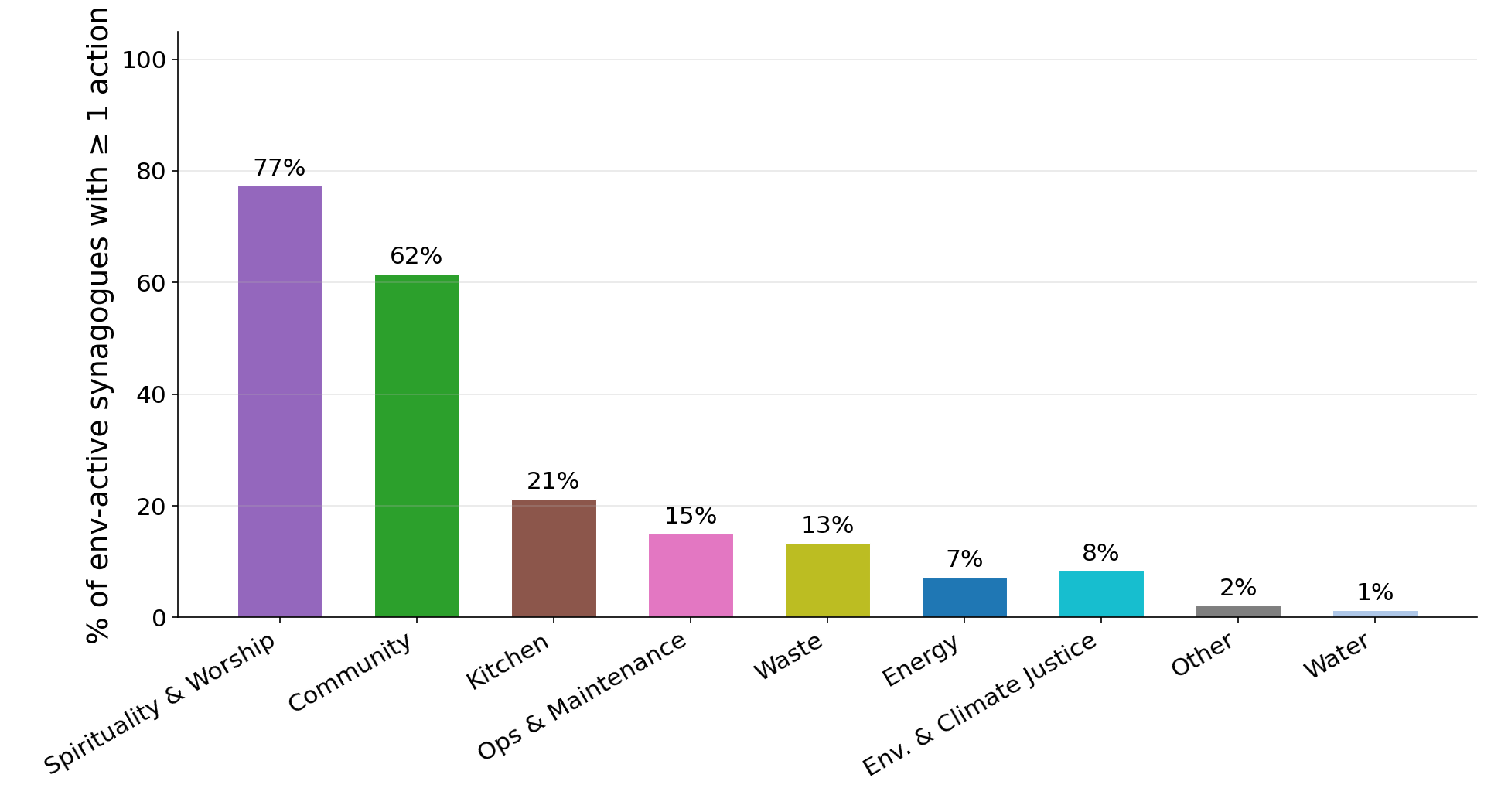}
            \caption{Percentage of environmentally active synagogues with at least one action in each environmental action category.}
        \label{fig:fig3a}
    \end{subfigure}
    \vspace{0.5cm}
    \begin{subfigure}{0.85\linewidth}
        \centering
    \includegraphics[width=\linewidth]{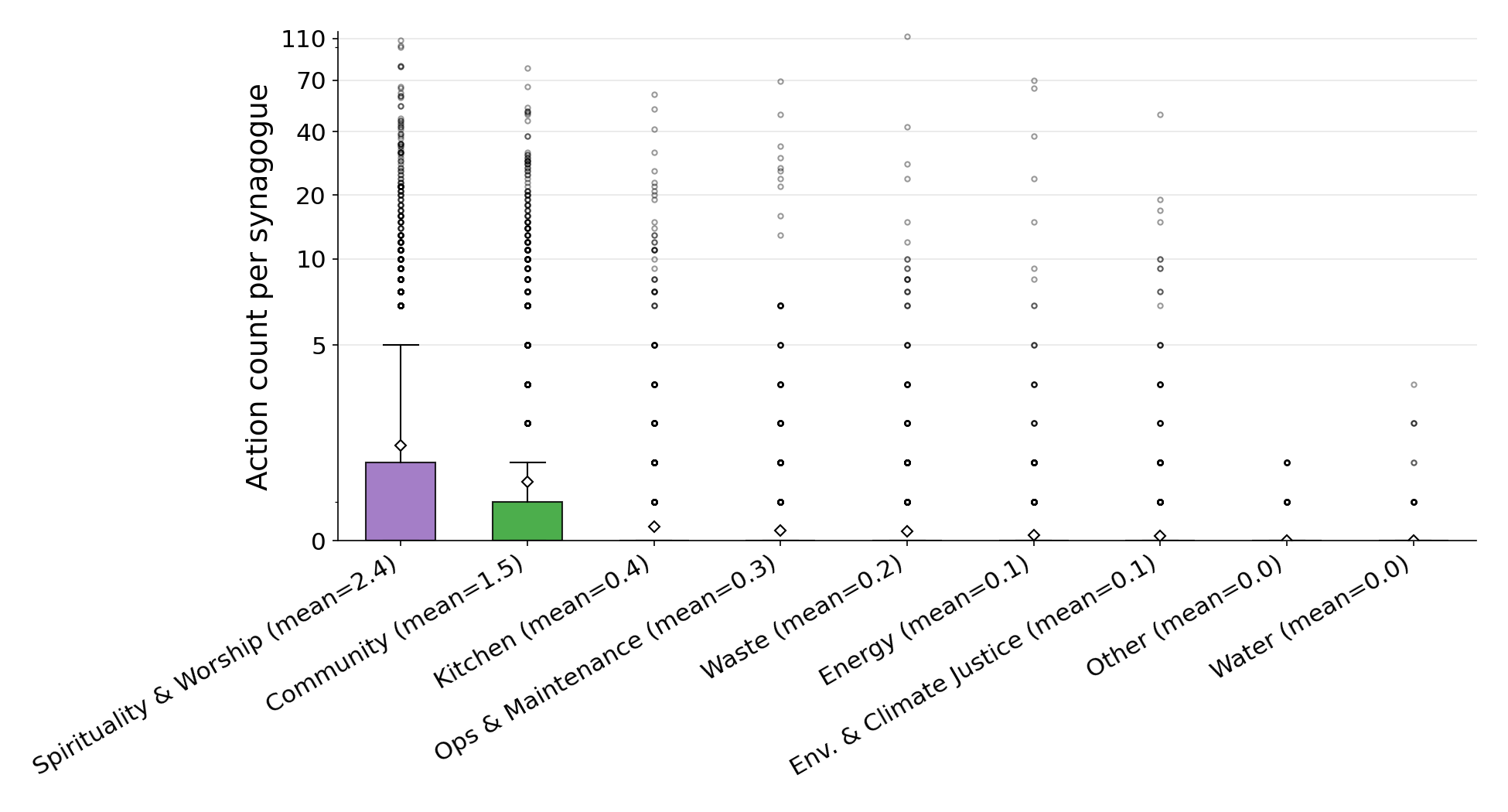}
\caption{Distribution of action counts per synagogue by environmental action category. The vertical axis uses a symmetric logarithmic scale to display the strongly right-skewed distributions. Diamonds show means.}
        \label{fig:fig3b}
    \end{subfigure}
\caption{Distribution of environmental actions by category.}
    \label{fig:fig3}
\end{figure}

Environmental action was concentrated in a small number of action categories (Figure \ref{fig:fig3a}). Among environmentally active congregations, 77\% had at least one identified Spirituality and Worship action and 62\% had at least one Community action. The remaining categories were substantially less prevalent, with 21\% of environmentally active congregations had at least one Kitchen action, 15\% Operations and Maintenance, 13\% Waste, 8\% Environmental and Climate Justice, 7\% Energy, 2\% Other, and 1\% Water.

The number of actions per congregation was also highest for Spirituality and Worship and Community with a mean of 6.6 actions and 4.4 actions, respectively. Together, these two categories accounted for 80\% of all identified environmental actions.

Action counts were strongly right skewed across categories, with most congregations having relatively few actions and a small number having substantially larger counts (Figure \ref{fig:fig3b}). This pattern was pronounced for Spirituality and Worship and Community, although long right tails were also evident in several less prevalent categories.

\subsection{Environmental Action by Denominational Affiliation}

\begin{figure}[H]
    \centering
    \includegraphics[width=0.85\linewidth]{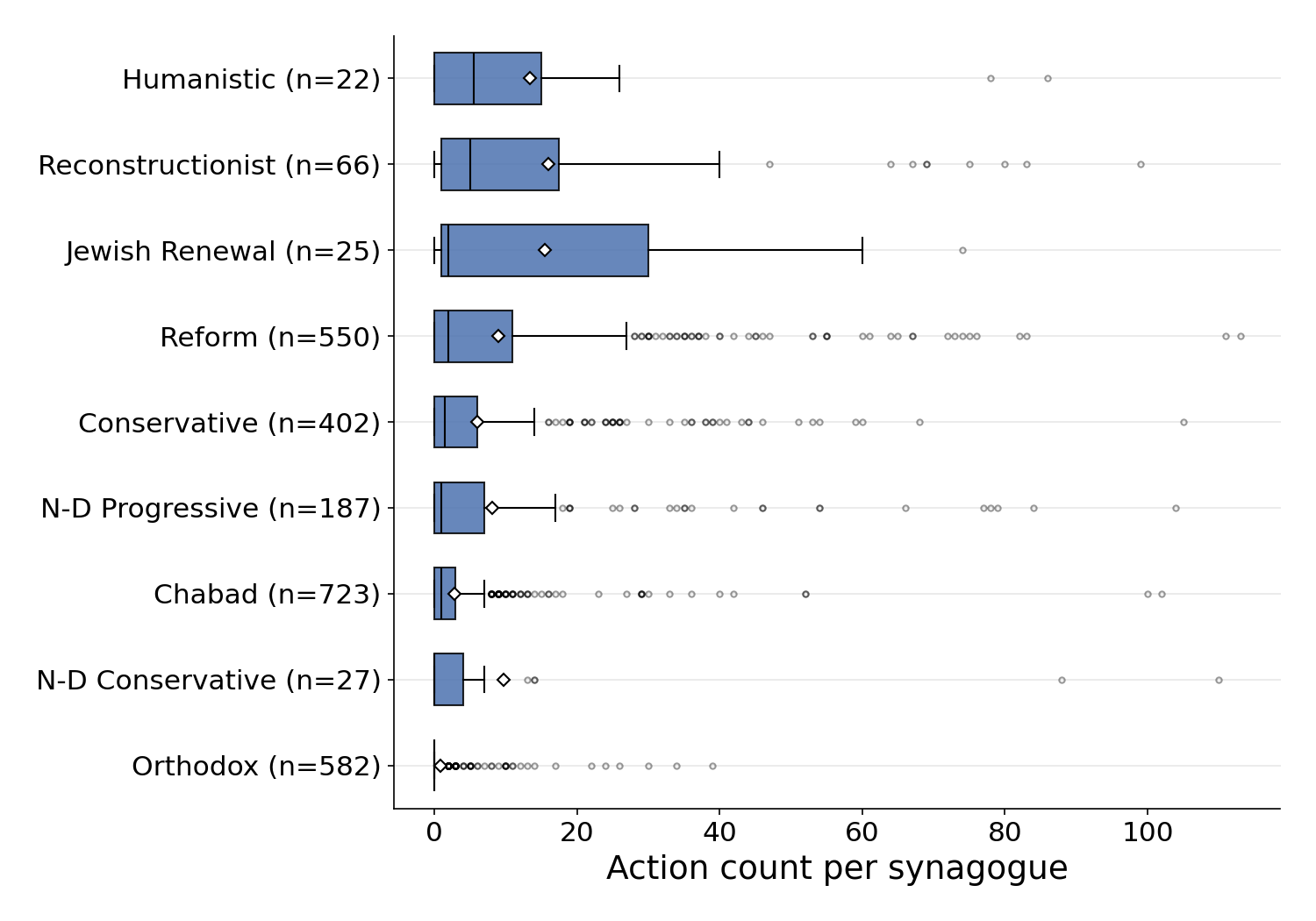}
    \caption{Distribution of environmental action counts by denominational affiliation among U.S. Jewish congregations. Diamonds show means and denominations are ordered according to medians.}
    \label{fig:fig4}
\end{figure}

Environmental action counts differed significantly across denominational groups (Kruskal-Wallis test $H(8) = 413.8$, $p < 0.001$, $n = 2584$) (Figure \ref{fig:fig4}). Congregations for which denominational affiliation could not be determined were excluded from this analysis. 

Post-hoc pairwise comparisons indicated Orthodox synagogues had statistically significantly lower environmental action counts than all other denominations. Chabad congregations had significantly lower action counts than Reconstructionist, Reform, Jewish Renewal, Non-Denominational Progressive and Conservative synagogues. Jewish Renewal and Reconstructionist synagogues both take significantly more action than Non-Denominational Conservative, Conservative and Non-Denominational Progressive synagogues in addition to the aforementioned significant differences with Chabad and Orthodox synagogues. 

\subsection{Environmental Framing}

\begin{figure}[ht]
    \centering
    \includegraphics[width=0.75\linewidth]{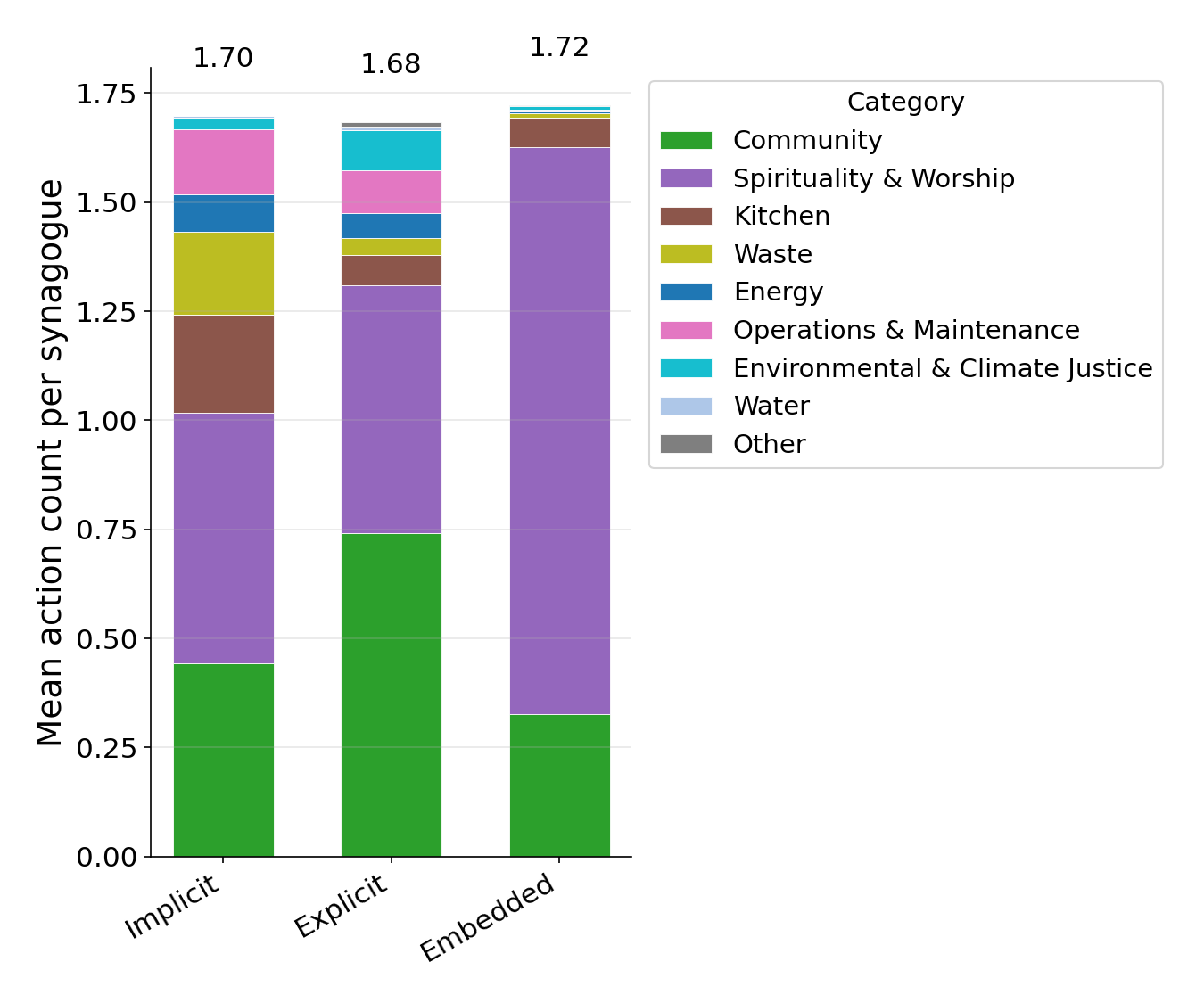}
    \caption{Mean environmental action count per congregation by environmental framing and action category. Stacked bars show the contribution of each environmental action category to the mean number of implicit, explicit, and embedded actions per congregation. Values above bars indicate the total mean action count for each framing type.}
    \label{fig:fig5}
\end{figure}

Mean environmental action count per synagogue was remarkably similar across the three framing types (Figure \ref{fig:fig5}), but their composition by action category differs substantially. According to a Friedman test, the differences in mean action count by framing was statistically significant ($\chi^2(2) = 53.8$, $p < 0.001$, $n = 2657$). The Nemenyi's test indicated that the distribution of actions framed as Explicit vs Embedded were statistically significantly different ($p = 0.045$). Implicit vs Explicit and Implicit vs Embedded were not significantly different ($p= 0.09, 0.90$)  Embedded actions were dominated by Spirituality and Worship, whereas explicit actions contained a larger contribution from Community actions. Implicit actions were more broadly distributed across action categories, including comparatively larger contributions from Kitchen, Waste, Energy, and Operations and Maintenance.

\begin{figure}[ht]
    \centering
    \includegraphics[width=0.95\linewidth]{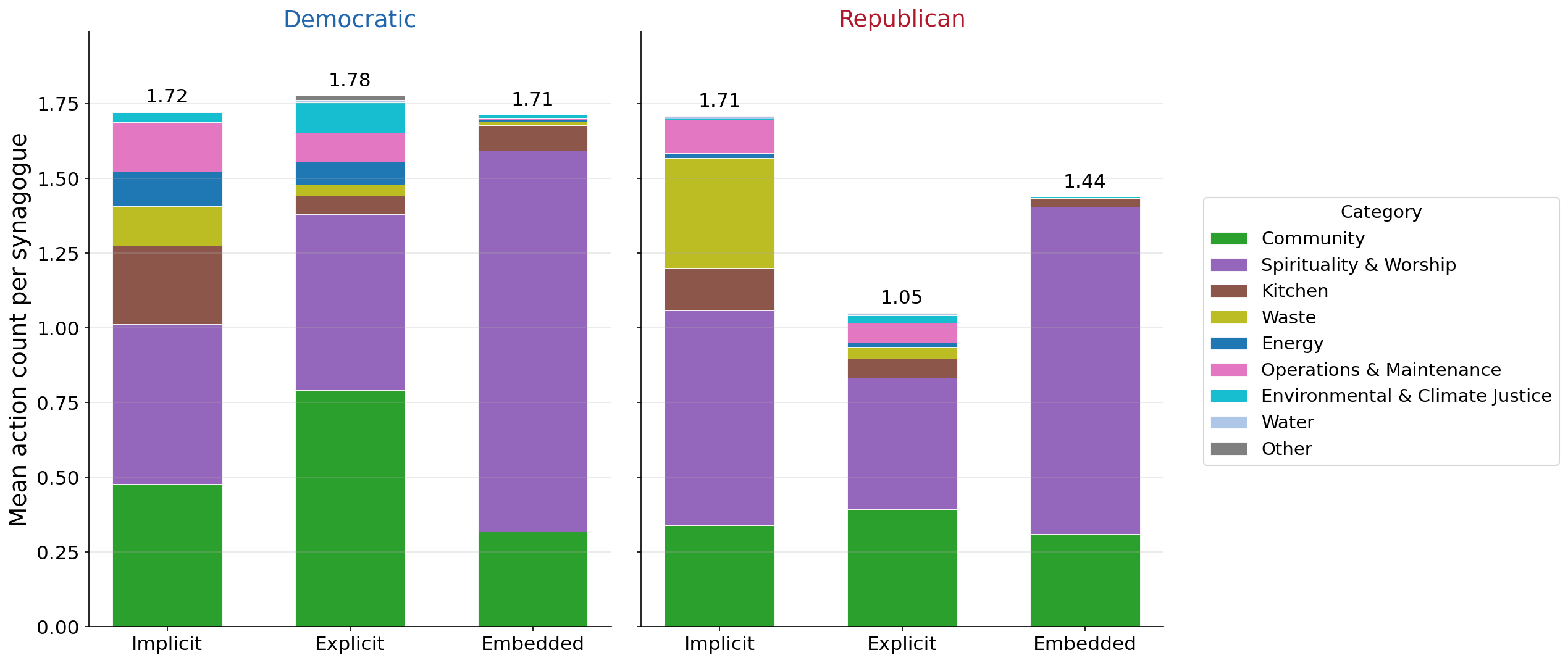}
    \caption{Mean environmental action counts per congregation by environmental framing and 2024 congressional district political context. Stacked bars show the contribution of each environmental action category to the mean number of implicit, explicit, and embedded actions among congregations located in districts won by Democratic and Republican candidates in the 2024 federal election. Values above bars indicate the total mean action count for each framing type.}
    \label{fig:fig6}
\end{figure}

Of the congregations included in the political context analysis, 1,856 were located in congressional districts won by Democratic candidates in the 2024 federal election and 680 were located in districts won by Republican candidates. Synagogues not definitively assigned to a district, such as those in Washington D.C. were not included. Mean environmental action counts differed descriptively across framing types and district political context (Figure \ref{fig:fig6}). Congregations in Democratic-won districts had mean implicit, explicit, and embedded action counts of 1.72, 1.78, and 1.71, respectively, compared with 1.71, 1.05, and 1.44 among congregations in Republican-won districts. The largest difference was observed for explicit environmental actions, whereas mean implicit action counts were nearly identical between the two groups.

Action count distributions differed significantly between Republican and Democratic district synagogues for both Explicit ($p=0.01$) and Implicit ($p<0.001$) actions.

\section{Discussion}

This study demonstrates that large-scale organizational web data can be transformed into structured measure of environmental action using automated web crawling and LLM-based information extraction. Applying this framework to Jewish congregations in the U.S.\ revealed environmental activity across a substantial proportion of congregations with accessible websites and considerable heterogeneity in the types, geographic distribution, denominational patterns, and framing of environmental action. More broadly, our findings highlight both the potential and the methodological challenges of using organizational websites as a source of environmental data.

A central contribution of this study is the comparison of alternative approaches for extracting environmental information from a large corpus of organizational websites. Keyword retrieval followed by LLM classification substantially reduced the amount of text requiring LLM evaluation and consequently had the lowest computational cost, but it also showed the lowest agreement with expert human classification. Semantic vector retrieval improved agreement while retaining much of the computational advantage of preliminary filtering. However, both retrieval approaches could exclude relevant content before LLM evaluation, with greater information loss under keyword retrieval. Direct LLM classification required greater computational resources but provided the broadest congregation-level coverage.

These findings illustrate an important distinction between information retrieval and information classification in environmental text analysis. A highly accurate classifier cannot recover relevant information that is excluded during an upstream retrieval step. Keyword-based retrieval is particularly vulnerable to this problem because organizations may describe environmentally relevant activities without using explicitly environmental terminology. Semantic vector retrieval reduces dependence on exact vocabulary but nevertheless requires the specification of a similarity threshold, creating a tradeoff between computational efficiency and retrieval recall. In applications where the objective is comprehensive detection of relatively rare environmental activities, our results suggest that reducing the candidate corpus before classification can introduce meaningful information loss.

The relatively strong agreement between direct LLM classification and expert review supports the use of general purpose LLMs for converting heterogeneous organizational text into structured environmental variables. However, LLM classification should not be interpreted as error-free measurement. Recent evaluations of LLM-as-a-judge approaches have demonstrated that model judgments can remain sensitive to task formulation and systematic biases \citep{chen2024}. Consequently, human validation remains an important component of LLM-based environmental data pipelines. Importantly, our results indicate that validation should encompass the complete information-extraction pipeline rather than the final classifier alone. Retrieval recall, classification agreement, and computational cost capture different aspects of pipeline performance and can lead to different conclusions about the preferred approach.

\subsection{Measuring Environmental Action from Web Data}

The dataset generated through this framework provides large-scale empirical evidence that environmental action is documented across U.S. Jewish congregations but is highly heterogeneous in both frequency and form. Environmental action counts were strongly right-skewed, with most congregations having relatively few identifiable actions and a smaller number exhibiting substantially greater documented environmental activity. This heterogeneity would be difficult to characterize using case studies or manually assembled examples and illustrates one advantage of national-scale web-based measurement.

Environmental action was particularly concentrated in the Spirituality and Worship and Community categories. Spirituality and Worship was both the most prevalent category and the category with the highest mean number of actions, followed by Community. These categories accounted for the large majority of identified environmental actions. This pattern is consistent with scholarship describing Jewish environmentalism as extending beyond conventional resource-management activities to include religious education, reinterpretation of Jewish texts, ritual practices, environmental ethics, and community engagement. Previous accounts of U.S. Jewish environmentalism have similarly emphasized the incorporation of ecological ideas into religious learning and practice alongside institutional greening and environmental advocacy \citep{tirosh-samuelsonJewishEnvironmentalismUnited2024}.

The prominence of Spirituality and Worship also illustrates why a broad operational definition of environmental action is important when studying religious organizations. Environmental engagement within congregations may not resemble environmental activity undertaken by governments, businesses, or explicitly environmental organizations. Religious environmentalism can instead be expressed through theological interpretation, education, ritual, and community practice. Restricting environmental measurement to conventional indicators such as renewable energy, recycling, or building efficiency would therefore omit an important component of environmental activity within faith communities.

At the same time, the prominence of spiritually embedded environmental activity presents a measurement challenge. Routine religious practices may be less likely to be documented on congregation websites than discrete programs, events, or initiatives. Consequently, embedded actions may be systematically underrepresented in web-derived data. The observed prevalence of embedded Spirituality and Worship actions should therefore be interpreted as evidence of documented activity rather than a complete census of religious environmental practice.

\subsection{Denominational Differences}

Environmental action counts differed substantially across denominational groups. Orthodox congregations generally exhibited fewer identified environmental actions than most other denominational groups, while Chabad congregations had fewer actions than Reconstructionist, Reform, and Conservative congregations. These results provide evidence of within-religion heterogeneity that is obscured when Judaism is treated as a single category in studies of religion and environmentalism.

The denominational differences are notable in light of previous scholarship suggesting broad institutional engagement with environmental issues across much of American Judaism. Tirosh-Samuelson documents environmental resolutions, theological engagement, and environmental programming across several major Jewish movements, while also emphasizing the diversity of American Jewish environmentalism \citep{tirosh-samuelsonJewishEnvironmentalismUnited2024}. Our congregation-level results suggest that broad denominational recognition of environmental issues does not necessarily translate into equivalent levels of documented environmental activity among local congregations.

These differences should not, however, be interpreted as demonstrating a causal effect of denomination. Denominational affiliation is associated with numerous characteristics that may also influence environmental engagement, including geography, congregational demographics, political context, institutional resources, and patterns of religious observance. The observed denominational differences therefore identify an important axis of heterogeneity but do not establish whether theology, institutional structure, membership composition, or other correlated characteristics are responsible for those differences.

\subsection{Environmental Framing}

The three forms of environmental framing occurred at similar overall mean frequencies but differed substantially in their composition. Embedded actions were dominated by Spirituality and Worship, whereas explicit actions contained a larger contribution from Community activities. Implicit actions were distributed more broadly across operational categories, including Kitchen, Waste, Energy, and Operations and Maintenance. These findings suggest that environmental framing is associated less with the overall quantity of environmental activity than with the types of activities through which environmental engagement is expressed.

This distinction has implications for environmental measurement. Activities with environmental consequences are not necessarily undertaken or communicated for explicitly environmental reasons. Energy conservation, food practices, waste reduction, maintenance decisions, or community programs may have environmental relevance while being motivated by financial, religious, social, or practical considerations. A measurement strategy restricted to actions explicitly described as “environmental” or “sustainable” would therefore capture only one component of organizational environmental behavior. Distinguishing explicit, implicit, and embedded actions provides a means of separating the environmental relevance of an activity from the motivations and narratives through which organizations communicate it.

\subsection{Political Context and Environmental Action}

Environmental framing also varied with the political context of the congressional district in which congregations were located. The largest descriptive difference occurred for explicitly framed environmental actions, which were less frequent among congregations located in districts won by Republican candidates than among congregations in districts won by Democratic candidates. Differences in implicit actions were substantially smaller, while embedded actions also showed variation between political contexts.

This pattern is consistent with extensive evidence that environmental and climate attitudes in the U.S. are politically polarized. Long-term analyses show that partisan identification has become one of the strongest correlates of environmental and climate attitudes in the U.S. \citep{smith2024, TarakeshwarEtAl2001,SherkatEllison2007, VincentnathanEtAl2016}. Research specifically examining religion and environmental attitudes likewise finds that political affiliation often explains more variation in climate attitudes than religious affiliation itself. Pew Research Center surveys have found substantial partisan differences in both concern about climate change and the frequency with which environmental issues are discussed in religious settings \citep{alper2022}.

The particularly large difference in explicit environmental actions may indicate that political context influences how congregations communicate environmental activity as well as whether particular activities occur. If some environmentally relevant practices are undertaken for non-environmental reasons, they may be less politically salient than actions explicitly framed around climate change, sustainability, or environmental protection. This possibility is consistent with the much smaller descriptive difference observed for implicit actions. However, congressional district voting is an ecological measure of political context and cannot be interpreted as the political affiliation of individual congregations or their members. Further research incorporating congregation-level political and demographic information would be required to distinguish contextual political effects from differences in congregation composition.

\subsection{Limitations}

Several limitations should be considered when interpreting these findings. First, the framework measures documented environmental action, not all environmental action undertaken by congregations. Congregations vary in the extent and manner in which they maintain websites, and some activities may never be published online. Website-based measurement may therefore systematically favor organizations with more extensive digital communication. Congregations without accessible websites could not be included in the environmental action analysis, introducing an additional potential source of selection bias. In addition, automated crawling successfully reached only 68\% of the 3,917 congregations for which a website was identified. Because of technical measures limiting automated access, some otherwise publicly accessible websites were not included in the corpus. These restrictions sometimes operated at the level of shared website platforms and therefore may have excluded groups of congregations rather than websites independently. If the use of particular website platforms or automated-access controls is associated with congregation characteristics that are themselves related to environmental engagement, the resulting corpus could systematically underrepresent some types of congregations.

Second, website content provides evidence that an organization reported or promoted an activity but does not independently verify that the activity occurred, its duration, or its environmental impact. The number of identified actions should therefore be interpreted as a measure of documented organizational environmental activity rather than a direct measure of environmental benefit. A congregation mentioning multiple environmental programs is not necessarily producing greater environmental benefits than a congregation undertaking one high-impact intervention.

Third, the unit of extracted text does not always correspond perfectly to a unique real-world action. The same activity may be described on multiple webpages, while a single passage may contain multiple related activities. Although the classification and aggregation procedures were designed to reduce duplication, action counts may partly reflect differences in website architecture and communication practices.

Fourth, LLM-based classification introduces measurement uncertainty. Human validation demonstrated useful agreement for environmental action detection and framing, as well as complete agreement for denominational classification within the prespecified validation sample, but none of these procedures should be assumed to be error-free. Although the validation sample size for denominational classification was determined to provide 80\% statistical power, the resulting estimate was based on only 20 congregations and therefore remains subject to sampling uncertainty. Model performance may also change across LLM versions, providers, prompts, or future website corpora, making preservation of prompts, model identifiers, source webpages, and validation procedures important for reproducibility. A further source of uncertainty relates to safety refusals. Gemini declined to classify 95 of the 1,583,017 (0.006\%) potential environmental action text chunks because of its safety mechanisms. Although these refusals represented a very small proportion of the corpus, the underlying criteria for individual refusals are not fully observable, making it difficult to determine whether the affected content differed systematically from successfully classified content.  

Fifth, the observed processing time reflects both computational requirements and constraints imposed by the Gemini developer API, which capped the number of calls available during processing at 350,000. As a result, classification of the direct LLM corpus required five days despite not requiring five continuous days of computation. API limits vary among LLM providers and service tiers and may change over time. Subsequent applications of this approach should therefore consider not only model performance and cost, but also throughput and provider-specific rate limits when selecting an LLM service. 

Finally, the present application is restricted to Jewish congregations in the U.S. and predominantly English-language web content. Both the environmental action taxonomy and the linguistic expressions used to describe environmental activity may differ across religious traditions, organizational types, countries, and languages. Application of the framework to other populations would therefore require domain-specific adaptation and renewed human validation.

\subsection{Implications and Future Directions}

Despite these limitations, this study demonstrates the potential of organizational websites as a scalable source of environmental data. Much environmental activity occurs at organizational scales that are poorly represented in conventional monitoring systems and administrative databases. Web-based information extraction provides a complementary approach for observing programs, commitments, practices, and forms of environmental engagement that would otherwise require costly surveys or manual data collection.

The framework is not limited to religious organizations. Similar pipelines could be applied to schools, universities, municipalities, nonprofit organizations, businesses, community organizations, and other institutions that communicate environmental initiatives through public websites. The combination of comprehensive web archiving, domain-specific environmental taxonomies, LLM-based classification, human validation, and geospatial linkage creates an extensible approach for converting unstructured organizational information into analysis-ready environmental data.

Future work could extend the framework in several directions. Longitudinal website collection could measure the emergence, persistence, and disappearance of environmental initiatives over time. Linking extracted actions to demographic, socioeconomic, political, climate, and environmental exposure data could help identify contextual predictors of organizational environmental engagement. Further methodological work could also evaluate alternative LLMs, retrieval strategies, uncertainty quantification, and active-learning approaches that selectively direct human review toward ambiguous classifications. Finally, validation against surveys, interviews, or direct observation could help establish how accurately web-documented environmental action represents activities occurring within organizations.

Overall, our findings demonstrate that LLM-assisted web-data extraction can provide a scalable means of measuring organizational environmental action while preserving the contextual richness of how those actions are communicated. For U.S. Jewish congregations, environmental engagement appears not as a single homogeneous behavior but as a heterogeneous combination of religious, community, operational, and explicitly environmental practices that vary across organizational and contextual characteristics. More broadly, the approach extends the range of data available for studying organizational environmental action beyond conventional structured sources to include activities documented in unstructured web content.


\section*{Acknowledgments}

This work was supported by the University of Toronto Data Sciences Institute Catalyst Grant DSI-CGY5R1P30 "A Data Science Framework for Quantifying Environmental Action in Faith Communities" and the Natural Science and Engineering Research Council of Canada.


\section*{Data and Code Availability}

Data to understand, reproduce and verify this analysis is available at \url{https://github.com/Reyno256/Quantifying-Environmental-Action}.


\bibliographystyle{abbrvnat}
\bibliography{refs}

\newpage
\begin{appendix}
\renewcommand{\thesection}{\Alph{section}}
\setcounter{section}{0}

\section{Appendix A}
\label{appendixA}

\subsection{Environmental Action Classification}
\label{appendixA1}
\begin{lstlisting}[language=Python, caption=Prompt]

"""You are a text classifier.

#Context: The data comes from {country} synagogue websites. Your task is to classify the environmental action discussed on the web page provided by the user.

# Task: Output the name of exactly one classification label from the list below. No explanation is needed. If the page does not contain an actual environmental action, return "N/A".

# Valid labels (return exactly one):
"Spirituality & Worship_Include symbols of nature in place of worship, gardens, or meditation areas"
"NA_Focus on environemntalism (stewardship, climate action, etc.) in worship for religious or community events (holidays, earth day, etc.)"
"NA_Incorporate environmental prayers/songs in worship"
"NA_Dedicate a worship service to environmentalism"
"NA_Offer retreats, prayers, or reflections to reconnect with nature"
"NA_Use environemtnally-friendly alternatives for items used in worship"
"Community_Promote and facilitate household energy audits and water conservation"
"NA_Assisted in local environmental cleanup, restoration projects or tree planting initiatives"
"NA_Encouraged car-pooling, public transit use, and/or bicycle use"
"NA_Encourage volunteering initiatives aimed at environmental responsibility, reducing individual carbon footprints, and taking climate action"
"NA_Resources with a nature and environmental stewardship focus are available in the faith communitys library, common area, or website"
"NA_Encourage members to adopt a greener lifestyle, to join community environmental projects, and to practice the 3 R's at home"
"NA_Planted a community garden for native pollinator-friendly plants and/or vegetables"
"NA_Bike racks and carpooling preferential parking spots have been installed on the faith community's property"
"NA_Hold or support a holiday or summer camp with an environmental stewardship/climate theme"
"NA_Hold educational sessions to educate people on environmental stewardship/climate action that link to the faith tradition and promote faith-based and local environmental organizations"
"NA_Undertake environmental projects in collaboration with other faith communities or local groups (e.g. planting pollinator gardens, planting trees, community spring clean-up, naturalization of schoolyards and sacred spaces etc.)"
"NA_Engage with local, provincial, or federal governments to support sustainable public policies"
"NA_A volunteer coordinator has been appointed to facilitate carpooling and to ensure all community members are aware of public transit routes to the facility"
"NA_Electric vehicle charging station(s) have been installed on the property or identified nearby"
"NA_Create and support opportunities for younger people to experience how people in their own and other communities are affected by the planetary crisis and how they can work for change"
"NA_A board committee focusing on environmental and sustainability issues has been established"
"Operations & Maintenance_Have a 'think twice before printing' policy and an active paper reduction and recycling policy"
"NA_Have a procurement policy for recycled and/or FSC certified paper (e.g. office, envelopes, bulletins, bathroom tissue, etc.)"
"NA_A sustainability coordinator and/or team has been created and are developing sustainable practices within the place of worship"
"NA_Biodegradable, non-toxic, phosphate-free cleaning products are purchased and used within the facility"
"NA_Make their own eco-friendly cleaning products using natural materials (e.g. baking soda, white vinegar, essential oils, etc.)"
"NA_Have a green maintenance plan including using alternatives to salt in the winter and/or seeking lot and yard maintenance providers who use green practices"
"Energy_Utility expenditures and energy consumption are tracked and documented"
"NA_Utility data is analyzed regularly to identify activities that cause high consumption. Patterns are identified and plans are in place to reduce consumption"
"NA_An energy audit has been completed by a trained professional auditor and they are working to resolve some of the identified issues"
"NA_Signage posted at all light switches reminds people to turn off lights when not in use"
"NA_Energy efficient lighting (e.g. LEDs) have been installed where possible. Dimmers, timers, and motion sensors are installed where possible"
"NA_Programmable thermostats are in use and winter/summer set-back temperatures are in place"
"NA_Heating/cooling system has a regular maintenance schedule, furnace filters are cleaned or replaced, and ducts are clean, where applicable"
"NA_High-efficiency heating/cooling or solar hot water system installed"
"NA_Ceiling fans are in regular use"
"NA_Usage of caulk and spray foam to air-seal the building"
"NA_Weather stripping on windows and doors has been upgraded"
"NA_Replaced their fossil fuel burning heating system with an air or ground source heat pump"
"NA_Insulation has been added to walls and roof"
"NA_Investigated the possibility of installing a renewable energy system of any kind in their facility"
"NA_Installed a renewable energy system in their facility"
"Water_Basic efficiency steps have been taken such as low-flow aerators on all faucets, posting signs advising people to limit water usage, and regular checks to ensure no leaky faucets or toilets"
"NA_Systems that limit total hot water usage such as a hot water heat recovery system or a pre-mixing of hot and cold water system have been installed"
"NA_Steps have been taken such as insulating hot water tanks and pipes"
"NA_Installed a high-efficiency hot water heater, such as an on-demand unit or a solar domestic hot water heating system"
"NA_Low-flush toilets and urinals are installed where possible throughout the facility"
"NA_A grey water system has been installed that allows the facility to reuse water for a second time or a large cistern has been installed to capture rainwater for interior building uses"
"NA_Exterior watering of the grounds only occurs in the morning or evening and is shut off when it is raining or rain is expected"
"NA_The grounds have been specifically landscaped with native pollinator plants that do not require much watering (xeriscaping) or installed a rain garden to capture runoff"
"NA_Rain barrels have been installed to collect roof run-off water for the purpose of watering the grounds"
"Waste_Blue bins can be found in every area. Signs are posted near the bins advising people as to what can and can't be recycled"
"NA_Recycling facilities are available at the faith community for dropping off items such as ink cartridges, cell phones, CFL bulbs, etc"
"NA_Items such as eye glasses and clothes are donated to the faith community for re-use or re-distribution"
"NA_Have a 'backyard' composting system for yard waste and kitchen waste"
"NA_Composts regularly through their municipal composting program"
"NA_Reusable or cloth bags are used while shopping at grocery stores to reduce plastic waste"
"NA_Fruits and vegetables are selected from a bin, rather than those on styrofoam trays and shrink-wrapped in plastic"
"NA_A waste reduction strategy has been planned and implemented. Track total waste to determine if targets are being met"
"Kitchen_Locally sourced food and drinks are used when possible"
"NA_Vegetarian, vegan, organic, and in-season food options are provided and served when possible"
"NA_Reusable, recyclable, and/or compostable dishes and cutlery are in regular use"
"NA_A policy of using washable dishes and cutlery, etc. during events and break times is in place"
"NA_Single-use dishes and cutlery have been completely eliminated from use"
"NA_Promotes fair trade through a purchasing policy and by offering fairly traded products for sale"
"NA_Community garden regularly provides fresh vegetables to events held at the facility or to local food programs"
"NA_Support local food initiatives through actions such as hosting a local farmers market or other types of food-related events"
"Environmental & Climate Justice_Studied its faith traditions recent statements on fossil fuel divestment"
"NA_Formed an environmental justice discussion group and climate justice themes have been incorporated into services and events and promoted across social media platforms"
"NA_Met with political leader(s) to urge them to honour the rights of Indigenous peoples to a clean and healthy environment and to take immediate action on behalf of Indigenous communities impacted by environmental pollution (land, water, air)"
"NA_Researched the traditional territory and treaty on which it is located and incorporates a statement of land acknowledgement into its services and events"
"NA_Hosted a workshop or discussion group concerning environmental issues being faced by Indigenous communities today, and committed to taking action"
"NA_Members have been provided with educational materials about fossil fuel divestment"
"NA_Hosts environmental/climate justice events (screenings, workshops, study sessions) and/or offers space for community groups to meet on these issues (online and in person)"
"NA_Reviewed their investments and agreed to divest from any fossil fuel investments OR to invest in renewable energy"
"NA_Met with political leader(s) to call for action on climate change and participating in a climate justice action"
"NA_Other"
"N/A"
"""
\end{lstlisting}

\subsection{Denomination Classification}
\label{appendixA2}
\begin{lstlisting}[language=Python, caption=Prompt]
    # Task: Output the name of exactly one classification label from the list below. No explanation is needed. If the page does not fit into one of those categories, return "Unkown".
 Classify denominations into one of 10 categories Reconstructionist, Reform, Jewish Renewal, Conservative, Non-Denominational Progressive, Humanistic, Non-Denominational Conservative, Modern Orthodox, Orthodox, Chabad or Unknown. 
\end{lstlisting}
\subsection{Framing Classification}
\label{appendixA3}
\begin{lstlisting}[language=Python, caption=Prompt]
    You are a text classifier.

# Context: The data comes from US synagogue websites. Each excerpt below describes
a confirmed environmental action (already classified into a category such as
"Energy_...", "Waste_...", "Spirituality & Worship_...", etc.). Your task is to
classify HOW the action is framed, using the typology from Baugh (2019) and
Caldwell, Probstein & Yoreh (2022):

- "Explicit": The action is presented with a STATED environmental,
  sustainability, climate, or nature-connection goal/purpose - the text gives a
  REASON tied to nature or the environment for doing the action (e.g. "to reduce
  our carbon footprint", "as part of our green/sustainability initiative", "to
  protect the environment", "to be more eco-friendly", "to connect with nature",
  "to explore our relationship to the natural world", "in honor of the earth").
  This applies even when the action ALSO happens in a religious/worship setting -
  if a stated reason references nature/environment/sustainability, choose
  "Explicit" over "Embedded".

- "Embedded": The action is part of regular religious practice, worship, or
  tradition, described using ONLY religious/spiritual language, with NO stated
  nature- or environment-related reason or purpose. Any environmental benefit is
  purely incidental - the text gives a religious reason (or no reason at all)
  for the action (e.g. a Tu B'Shvat seder, planting trees as a mitzvah or Jewish
  tradition, a "Tree of Life" symbol/name, blessings, land acknowledgement as
  part of services). If the text mentions nature/gardens/trees ONLY as religious
  symbolism or as the NAME/TITLE of a program, with no stated nature-connection
  purpose, choose "Embedded".

- "Implicit": The action clearly has an environmental effect or benefit
  (recycling, composting, energy-efficient lighting, water conservation, etc.),
  but is described matter-of-factly as a practical or operational thing the
  congregation does - WITHOUT being framed as either explicitly environmental
  or as part of religious practice/worship.

# Decision priority (apply in this order):
1. Does the text state a reason connecting the action to nature, the
   environment, climate, or sustainability - even briefly, even alongside
   religious language? -> "Explicit"
2. Otherwise, is the action framed in religious/spiritual/traditional terms (or
   merely named/titled using nature imagery), with no such stated reason?
   -> "Embedded"
3. Otherwise (practical/operational description, no religious or
   environmental framing) -> "Implicit"

# Task: Output the name of exactly one label from the list below. No explanation
is needed.

# Valid labels (return exactly one):
"Explicit"
"Embedded"
"Implicit"
\end{lstlisting}
\section{Appendix B}
\label{appendixB}
\subsection{Environmental Action Taxonomy}

\begin{longtable}{>{\raggedright\arraybackslash}p{2.4cm} >{\raggedright\arraybackslash}p{11.5cm}}
\caption{Environmental Action Taxonomy}
\label{tab:action_taxonomy} \\
\hline
\textbf{Category} & \textbf{Sub-Category} \\
\hline
\endfirsthead
\hline
\textbf{Category} & \textbf{Sub-Category} \\
\hline
\endhead
\hline
\endfoot
\hline
\endlastfoot
\multirow{6}{2.4cm}{\textbf{Spirituality \& Worship}}
 & Include symbols of nature in place of worship, gardens, or meditation areas \\
 & Focus on environmentalism (stewardship, climate action, etc.) in worship for religious or community events (holidays, earth day, etc.) \\
 & Incorporate environmental prayers/songs in worship \\
 & Dedicate a worship service to environmentalism \\
 & Offer retreats, prayers, or reflections to reconnect with nature \\
 & Use environmentally-friendly alternatives for items used in worship \\
\hline
\multirow{16}{2.4cm}{\textbf{Community}}
 & Promote and facilitate household energy audits and water conservation \\
 & Assisted in local environmental cleanup, restoration projects or tree planting initiatives \\
 & Encouraged car-pooling, public transit use, and/or bicycle use \\
 & Encourage volunteering initiatives aimed at environmental responsibility, reducing individual carbon footprints, and taking climate action \\
 & Resources with a nature and environmental stewardship focus are available in the faith communitys library, common area, or website \\
 & Encourage members to adopt a greener lifestyle, to join community environmental projects, and to practice the 3 R's at home \\
 & Planted a community garden for native pollinator-friendly plants and/or vegetables \\
 & Bike racks and carpooling preferential parking spots have been installed on the faith community's property \\
 & Hold or support a holiday or summer camp with an environmental stewardship/climate theme \\
 & Hold educational sessions to educate people on environmental stewardship/climate action that link to the faith tradition and promote faith-based and local environmental organizations \\
 & Undertake environmental projects in collaboration with other faith communities or local groups (e.g. planting pollinator gardens, planting trees, community spring clean-up, naturalization of schoolyards and sacred spaces etc.) \\
 & Engage with local, provincial, or federal governments to support sustainable public policies \\
 & A volunteer coordinator has been appointed to facilitate carpooling and to ensure all community members are aware of public transit routes to the facility \\
 & Electric vehicle charging station(s) have been installed on the property or identified nearby \\
 & Create and support opportunities for younger people to experience how people in their own and other communities are affected by the planetary crisis and how they can work for change \\
 & A board committee focusing on environmental and sustainability issues has been established \\
\hline
\multirow{6}{2.4cm}{\textbf{Operations \& Maintenance}}
 & Have a `think twice before printing' policy and an active paper reduction and recycling policy \\
 & Have a procurement policy for recycled and/or FSC certified paper (e.g. office, envelopes, bulletins, bathroom tissue, etc.) \\
 & A sustainability coordinator and/or team has been created and are developing sustainable practices within the place of worship \\
 & Biodegradable, non-toxic, phosphate-free cleaning products are purchased and used within the facility \\
 & Make their own eco-friendly cleaning products using natural materials (e.g. baking soda, white vinegar, essential oils, etc.) \\
 & Have a green maintenance plan including using alternatives to salt in the winter and/or seeking lot and yard maintenance providers who use green practices \\
\hline
\multirow{15}{2.4cm}{\textbf{Energy}}
 & Utility expenditures and energy consumption are tracked and documented \\
 & Utility data is analyzed regularly to identify activities that cause high consumption. Patterns are identified and plans are in place to reduce consumption \\
 & An energy audit has been completed by a trained professional auditor and they are working to resolve some of the identified issues \\
 & Signage posted at all light switches reminds people to turn off lights when not in use \\
 & Energy efficient lighting (e.g. LEDs) have been installed where possible. Dimmers, timers, and motion sensors are installed where possible \\
 & Programmable thermostats are in use and winter/summer set-back temperatures are in place \\
 & Heating/cooling system has a regular maintenance schedule, furnace filters are cleaned or replaced, and ducts are clean, where applicable \\
 & High-efficiency heating/cooling or solar hot water system installed \\
 & Ceiling fans are in regular use \\
 & Usage of caulk and spray foam to air-seal the building \\
 & Weather stripping on windows and doors has been upgraded \\
 & Replaced their fossil fuel burning heating system with an air or ground source heat pump \\
 & Insulation has been added to walls and roof \\
 & Investigated the possibility of installing a renewable energy system of any kind in their facility \\
 & Installed a renewable energy system in their facility \\
\hline
\multirow{9}{2.4cm}{\textbf{Water}}
 & Basic efficiency steps have been taken such as low-flow aerators on all faucets, posting signs advising people to limit water usage, and regular checks to ensure no leaky faucets or toilets \\
 & Systems that limit total hot water usage such as a hot water heat recovery system or a pre-mixing of hot and cold water system have been installed \\
 & Steps have been taken such as insulating hot water tanks and pipes \\
 & Installed a high-efficiency hot water heater, such as an on-demand unit or a solar domestic hot water heating system \\
 & Low-flush toilets and urinals are installed where possible throughout the facility \\
 & A grey water system has been installed that allows the facility to reuse water for a second time or a large cistern has been installed to capture rainwater for interior building uses \\
 & Exterior watering of the grounds only occurs in the morning or evening and is shut off when it is raining or rain is expected \\
 & The grounds have been specifically landscaped with native pollinator plants that do not require much watering (xeriscaping) or installed a rain garden to capture runoff \\
 & Rain barrels have been installed to collect roof run-off water for the purpose of watering the grounds \\
\hline
\multirow{8}{2.4cm}{\textbf{Waste}}
 & Blue bins can be found in every area. Signs are posted near the bins advising people as to what can and can't be recycled \\
 & Recycling facilities are available at the faith community for dropping off items such as ink cartridges, cell phones, CFL bulbs, etc \\
 & Items such as eye glasses and clothes are donated to the faith community for re-use or re-distribution \\
 & Have a `backyard' composting system for yard waste and kitchen waste \\
 & Composts regularly through their municipal composting program \\
 & Reusable or cloth bags are used while shopping at grocery stores to reduce plastic waste \\
 & Fruits and vegetables are selected from a bin, rather than those on styrofoam trays and shrink-wrapped in plastic \\
 & A waste reduction strategy has been planned and implemented. Track total waste to determine if targets are being met \\
\hline
\multirow{8}{2.4cm}{\textbf{Kitchen}}
 & Locally sourced food and drinks are used when possible \\
 & Vegetarian, vegan, organic, and in-season food options are provided and served when possible \\
 & Reusable, recyclable, and/or compostable dishes and cutlery are in regular use \\
 & A policy of using washable dishes and cutlery, etc. during events and break times is in place \\
 & Single-use dishes and cutlery have been completely eliminated from use \\
 & Promotes fair trade through a purchasing policy and by offering fairly traded products for sale \\
 & Community garden regularly provides fresh vegetables to events held at the facility or to local food programs \\
 & Support local food initiatives through actions such as hosting a local farmers market or other types of food-related events \\
\hline
\multirow{9}{2.4cm}{\textbf{Environmental \& Climate Justice}}
 & Studied its faith tradition's recent statements on fossil fuel divestment \\
 & Formed an environmental justice discussion group and climate justice themes have been incorporated into services and events and promoted across social media platforms \\
 & Met with political leader(s) to urge them to honour the rights of Indigenous peoples to a clean and healthy environment and to take immediate action on behalf of Indigenous communities impacted by environmental pollution (land, water, air) \\
 & Researched the traditional territory and treaty on which it is located and incorporates a statement of land acknowledgement into its services and events \\
 & Hosted a workshop or discussion group concerning environmental issues being faced by Indigenous communities today, and committed to taking action \\
 & Members have been provided with educational materials about fossil fuel divestment \\
 & Hosts environmental/climate justice events (screenings, workshops, study sessions) and/or offers space for community groups to meet on these issues (online and in person) \\
 & Reviewed their investments and agreed to divest from any fossil fuel investments OR to invest in renewable energy \\
 & Met with political leader(s) to call for action on climate change and participating in a climate justice action \\
\hline
\textbf{Other} & \\
\end{longtable}

\end{appendix}

\end{document}